\documentclass[lettersize,journal]{IEEEtran}
\usepackage{amsmath,amsfonts}
\usepackage{booktabs}  
\usepackage{tabularx}
\usepackage{algorithmic}
\usepackage{algorithm}
\usepackage{array}
\usepackage[caption=false,font=normalsize,labelfont=sf,textfont=sf]{subfig}
\usepackage{textcomp}
\usepackage{stfloats}
\usepackage{url}
\usepackage{verbatim}
\usepackage{graphicx}
\usepackage{cite}
\usepackage{float}

\begin{document}

\title{DiffColor: Toward High Fidelity Text-Guided Image Colorization with Diffusion Models}

\author{Jianxin~Lin, 
Peng~Xiao, 
Yijun~Wang, 
Rongju Zhang, 
Xiangxiang Zeng~\IEEEmembership{Senior~Member,~IEEE}
\thanks{Jianxin Lin, Peng Xiao, Yijun Wang, Rongju Zhang, and Xiangxiang Zeng are with the College of Computer Science and Electronic Engineering, Hunan University, Changsha 410082, China (e-mail:	linjianxin@hnu.edu.cn; napping@hnu.edu.cn; wyjun@hnu.edu.cn; 1251553178@qq.com; xzeng@foxmail.com).}}

\markboth{}%
{Shell \MakeLowercase{\textit{et al.}}: A Sample Article Using IEEEtran.cls for IEEE Journals}


\maketitle

\begin{abstract}
Recent data-driven image colorization methods have enabled automatic or reference-based colorization, while still suffering from unsatisfactory and inaccurate object-level color control. To address these issues, we propose a new method called DiffColor that leverages the power of pre-trained diffusion models to recover vivid colors conditioned on a prompt text, without any additional inputs. DiffColor mainly contains two stages: colorization with generative color prior and in-context controllable colorization. Specifically, we first fine-tune a pre-trained text-to-image model to generate colorized images using a CLIP-based contrastive loss. Then we try to obtain an optimized text embedding aligning the colorized image and the  text prompt, and a fine-tuned diffusion model enabling   high-quality image reconstruction. Our method can produce vivid and diverse colors with a few iterations, and keep the structure and background intact while having colors well-aligned with the target language guidance. Moreover, our method allows for in-context colorization, i.e., producing different colorization results by modifying prompt texts without any fine-tuning, and can achieve object-level controllable colorization results. Extensive experiments and user studies demonstrate that DiffColor outperforms previous works in terms of visual quality, color fidelity, and diversity of colorization options. 
\end{abstract}


\section{Introduction}
\begin{figure*}
  \includegraphics[width=1\textwidth]{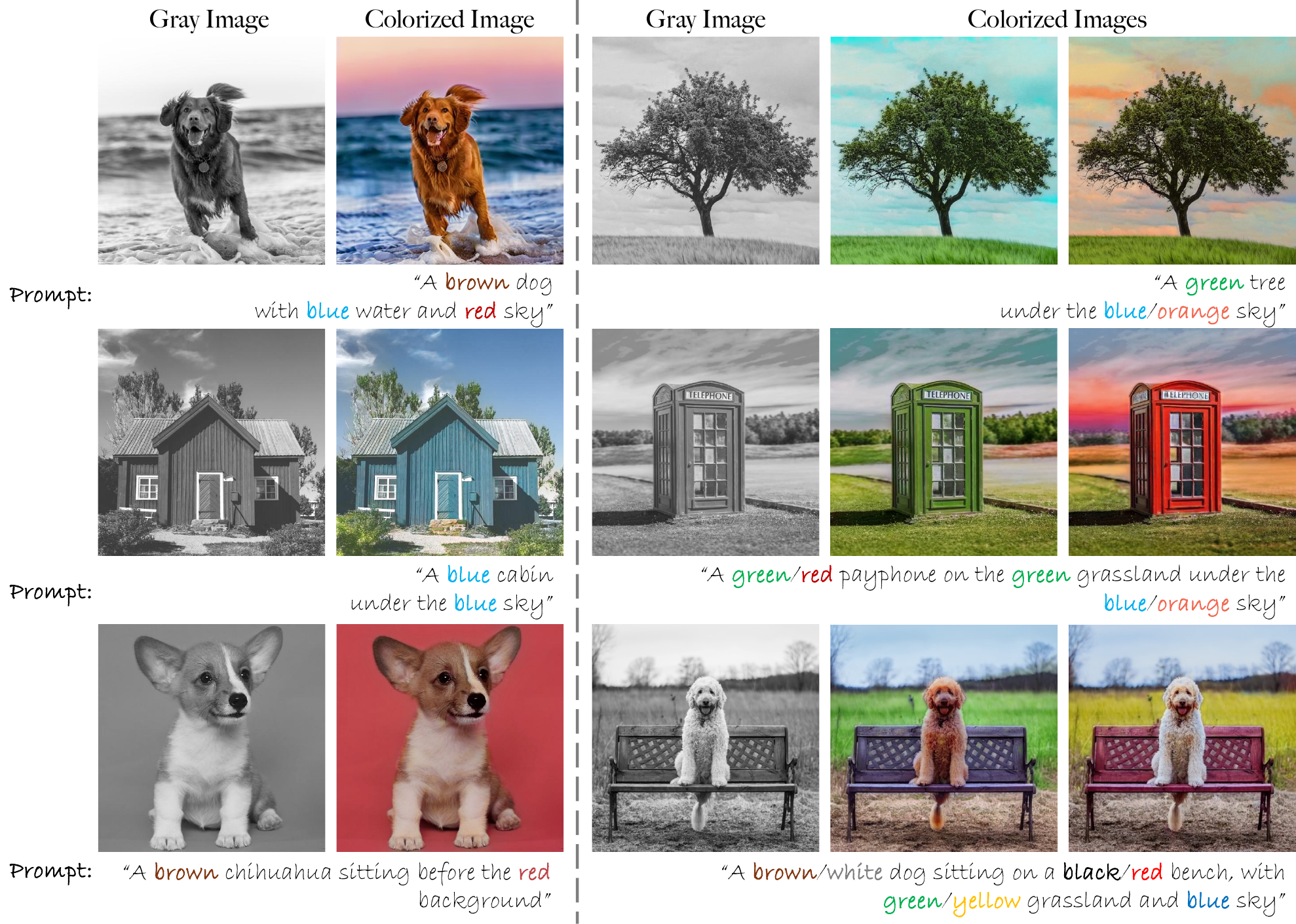}
  \caption{\textbf{Text-Guided Image Colorization.} Given \emph{a grayscale image}  and \emph{a target text prompt} with color description, our proposed approach DiffColor can produce \emph{high-quality} colorization results and realize \emph{object-level controllable} colorization. }
  \label{fig:title_page}
\end{figure*}

Image colorization, the process of adding colors to black-and-white images, has gained significant attention in recent years, with applications in various fields such as photography, advertising, and the film industry. However, colorization is inherently an ill-posed problem due to the fact that it requires estimating missing color channels from only one grayscale value, making it a challenging task.  Traditional colorization methods typically require user intervention, such as providing color scribbles \cite{huang2005adaptive, levin2004colorization, yatziv2006fast} or reference images \cite{ironi2005colorization,gupta2012image,charpiat2008automatic} to obtain satisfactory results. With the advent of deep learning, various techniques have been developed that leverage deep neural networks and large-scale datasets such as ImageNet \cite{deng2009imagenet} or COCO-Stuff \cite{caesar2018coco} to learn colorization in an end-to-end fashion \cite{guadarrama2017pixcolor, bahng2018coloring, iizuka2016let, lei2019fully, su2020instance}.  



Numerous existing colorization methods \cite{guadarrama2017pixcolor, iizuka2016let, isola2017image} employ pixel-level regression models to learn color distribution priors from large-scale training data. However, these methods tend to produce brownish average colors for objects with inherent color ambiguity, such as clothes, houses, other man-made objects, and so on. To enhance the variety of produced colors, other researchers \cite{larsson2016learning, zhang2016colorful, zhao2018pixel} suggested predicting per-pixel color distributions instead of a single color. Additionally, reference-based approaches \cite{wu2021towards, camilo2022super, he2018deep} propose to transfer color statistics from a reference image to a grayscale image using correspondences. However, their outputs often exhibit visual distortions such as irregular and inconsistent colors. Other works \cite{wang2022palgan,isola2017image,wu2021towards} also resort to Generative Adversarial Networks (GANs) \cite{goodfellow2020generative} to encourage generated chroma distributions to be indistinguishable from real-life color images. 

Recently, diffusion-based models like denoising diffusion probabilistic models (DDPM) \cite{ho2020denoising, sohl2015deep} and score-based generative models \cite{song2019generative, song2020score} have made significant progress in image generation tasks \cite{ho2020denoising, song2020score}. The latest studies \cite{dhariwal2021diffusion, song2020score} have demonstrated even higher quality of image synthesis compared to other generative models such as variational autoencoders (VAEs) \cite{kingma2013auto, van2017neural, razavi2019generating}, flows \cite{dinh2016density, kingma2018glow, rezende2015variational}, auto-regressive models \cite{menick2018generating, van2016pixel}, and generative adversarial networks (GANs) \cite{brock2018large, goodfellow2020generative, karras2019style}. Moreover, a recent denoising diffusion implicit model (DDIM) \cite{song2020denoising} has further improved the sampling efficiency and enabled almost perfect inversion \cite{dhariwal2021diffusion}.

In this work, we propose a novel approach, DiffColor, for high-fidelity text-guided image colorization using diffusion models, which leverages the power of text descriptions to provide accurate and diverse colorization results. Text-guided image colorization has the advantage of being able to incorporate prior knowledge about the object or scene being colorized, making it particularly useful in domains where such prior knowledge is available. DiffColor aims to recover vivid and diverse colors by using pre-trained diffusion models \cite{rombach2022high,saharia2022photorealistic}, to incorporate text descriptions into the colorization process.

DiffColor mainly contains two stages: colorization with generative color prior and in-context controllable colorization. Specifically, we first convert an input grayscale image into latent noises through forward diffusion. We fine-tune the score function in the
reverse diffusion process, conditioned on the context description text embedding, to enrich the generated image's colors by using a contrastive loss. The contrastive loss is built by taking a colorized image and context text as a pair of positive samples and other anti-color texts as negative samples, incorporating more accurate generative color priors into colorization results. Then, we try to obtain an optimized text embedding aligning the colorized image and the  text prompt, and a fine-tuned diffusion model enabling  high-quality image reconstruction. Finally, we can achieve two goals:  
1) our method allows for in-context colorization, in which, unlike existing text-driven image manipulation methods, DiffColor can generate different colorization results by simply modifying prompt texts without any fine-tuning; 2) our method can produce object-level controllable colorization results.

We demonstrate the effectiveness of DiffColor through extensive experiments and user studies. The results show that our method produces high-quality images that closely resemble the input image and align well with the target text, and outperforms previous state-of-the-art methods in terms of colorization quality and diversity. Furthermore, we conducted an ablation study to evaluate the effect of each element of our method.

Our contributions can be summarized as follows:
\begin{itemize}
    \item Development of a new framework: The paper presents DiffColor which utilizes text-guided diffusion models for the task of image colorization. This framework is designed to achieve high fidelity and controllable colorization results compared to existing methods.
    \item Incorporation of a novel color contrastive loss as guidance: The proposed framework incorporates a color contrastive loss by leveraging both context text and other anti-color texts, bringing higher quality colorization results.   
    \item Introducing a two-stage color refinement process:  we further propose to refine the initial colorized image reconstruction and text embedding,  resulting in in-context and object-level controllable image colorization.

    \item Evaluation and comparison: The paper presents extensive experiments and evaluations of the proposed framework, demonstrating its superiority over existing methods in terms of colorization quality, diversity, and controllability. These evaluations also highlight the effectiveness of the text-guided approach and its potential for real-world applications.
\end{itemize}


\section{Related Works}

\noindent\textbf{Learning-based colorization.}
Learning-based colorization methods automatically colorize grayscale images with large-scale training data and end-to-end learning models. This line of work mainly addresses the two key issues of color semantics and multi-modality in image colorization. 
PalGAN \cite{wang2022palgan} predicts the pixel colors in a coarse-to-fine paradigm, decomposing colorization to palette estimation and pixel-wise assignment. Specifically, to build better semantic representation, Iizuka et al. \cite{iizuka2016let} and Zhao et al. \cite{zhao2018pixel} present a two-branch architecture that jointly learns and fuse local image features and global priors (e.g., semantic labels).  Su et al. \cite{su2020instance} learn object-level semantics by training on the cropped object images. To address the challenge of multi-modality, several studies \cite{larsson2016learning, zhang2016colorful} have suggested representing color prediction as pixel-level color classification. These approaches enable the assignment of multiple colors to each pixel based on the posterior probability.


To tackle the uncertainty and diversity problems of colorization, reference-based methods attempt to transfer the color statistics from a reference image to the input grayscale image \cite{charpiat2008automatic, gupta2012image, liu2008intrinsic,wu2021towards, camilo2022super, he2018deep}, which also enabled automatic colorization. The correspondences between the input and reference images are computed based on low-level similarity measures at pixel \cite{liu2008intrinsic, welsh2002transferring}, semantic segments \cite{charpiat2008automatic, ironi2005colorization}, or super-pixel levels \cite{chia2011semantic, gupta2012image}. These methods highly rely on the time-consuming procedure of reference image retrieving as well as manual annotations of image regions \cite{chia2011semantic, ironi2005colorization}, and often suffer from unsatisfactory and inaccurate object-level color control. Then Wu et al. \cite{wu2021towards} investigate the integration of a generative color prior learned from a pretrained BigGAN \cite{brock2018large} to improve the ability of a deep learning model to produce diverse colored results.



These methods were insufficient in providing accurate semantic guidance, such as natural language, for coloring. How to enhance the semantic modeling capability of context, and how to better incorporate semantic priors into the coloring process, are issues that need to be addressed.

\noindent\textbf{Text-driven image synthesis.} Text-to-image synthesis has been an active research area in the field of generative models, with numerous works focusing on generating realistic images from textual descriptions. Initial approaches relied on RNN-based methods \cite{mansimov2015generating}, which were later replaced by generative adversarial networks (GANs) \cite{brock2018large, goodfellow2020generative, karras2019style} that yielded improved results \cite{reed2016generative}. Subsequent enhancements to GAN-based models included multi-stage architectures and attention mechanisms \cite{xu2018attngan, zhang2017stackgan, zhang2018stackgan++}. DALL-E \cite{ramesh2021zero} introduced a two-stage approach that used a discrete VAE and transformer to model the joint distribution of text and image tokens without relying on GANs. New solutions to the problem of text-guided image synthesis were introduced with the recent development of diffusion models \cite{ho2020denoising, rombach2022high, sohl2015deep, song2020denoising, song2019generative}, which have demonstrated impressive results. However, these methods lack the ability to edit parts of a real image while preserving the rest. As a result, recent works have focused on taking advantage of these models rather than training a large-scale text-to-image model from scratch.


\noindent\textbf{Text-driven image manipulation.} 
Recently, text-driven image manipulation has achieved significant progress using GANs \cite{brock2018large, goodfellow2020generative, karras2021alias, karras2019style, karras2020analyzing}, which are known for their high-quality generation, further with CLIP \cite{radford2021learning}, which consists of a semantically rich joint image-text representation, pre-trained over millions of text-image pairs. Works that combined these components \cite{abdal2022clip2stylegan, gal2022stylegan, patashnik2021styleclip, xia2021tedigan} produced highly realistic manipulations using text only.
To obtain more expressive generation capabilities, recent approaches often use pre-trained  diffusion models for image editing. DiffusionCLIP \cite{kim2022diffusionclip} uses the CLIP model to obtain gradients for image manipulation, and achieves remarkable results in style transfer.  DreamBooth \cite{ruiz2022dreambooth} fine-tune the complete diffusion model by utilizing a small set of personalized images to create images of the same object in a new environment. Imagic \cite{kawar2022imagic} accepts a single image and a simple text prompt describing the desired edit, and aims to apply this edit while preserving a maximal amount of details from the image. 
L-code \cite{weng2022code} and Unicolor \cite{huang2022unicolor} also proposed to colorize images with text prompts. However, unlike L-code and Unicolor require large-scale training, we are the first work proposing to leverage the powerful text-to-image generation model, i.e., a pre-trained stable diffusion model, for text-guided image colorization.

In this work, we provide a novel approach for text-guided image colorization which can achieve fidelity and text alignment simultaneously, performing high-quality colorization globally and locally in one single image. The resulting image outputs align well with the target text while preserving the overall background, structure, and composition of the original image.

\begin{figure*}[ht]
	\centering
	\includegraphics[scale=0.53]{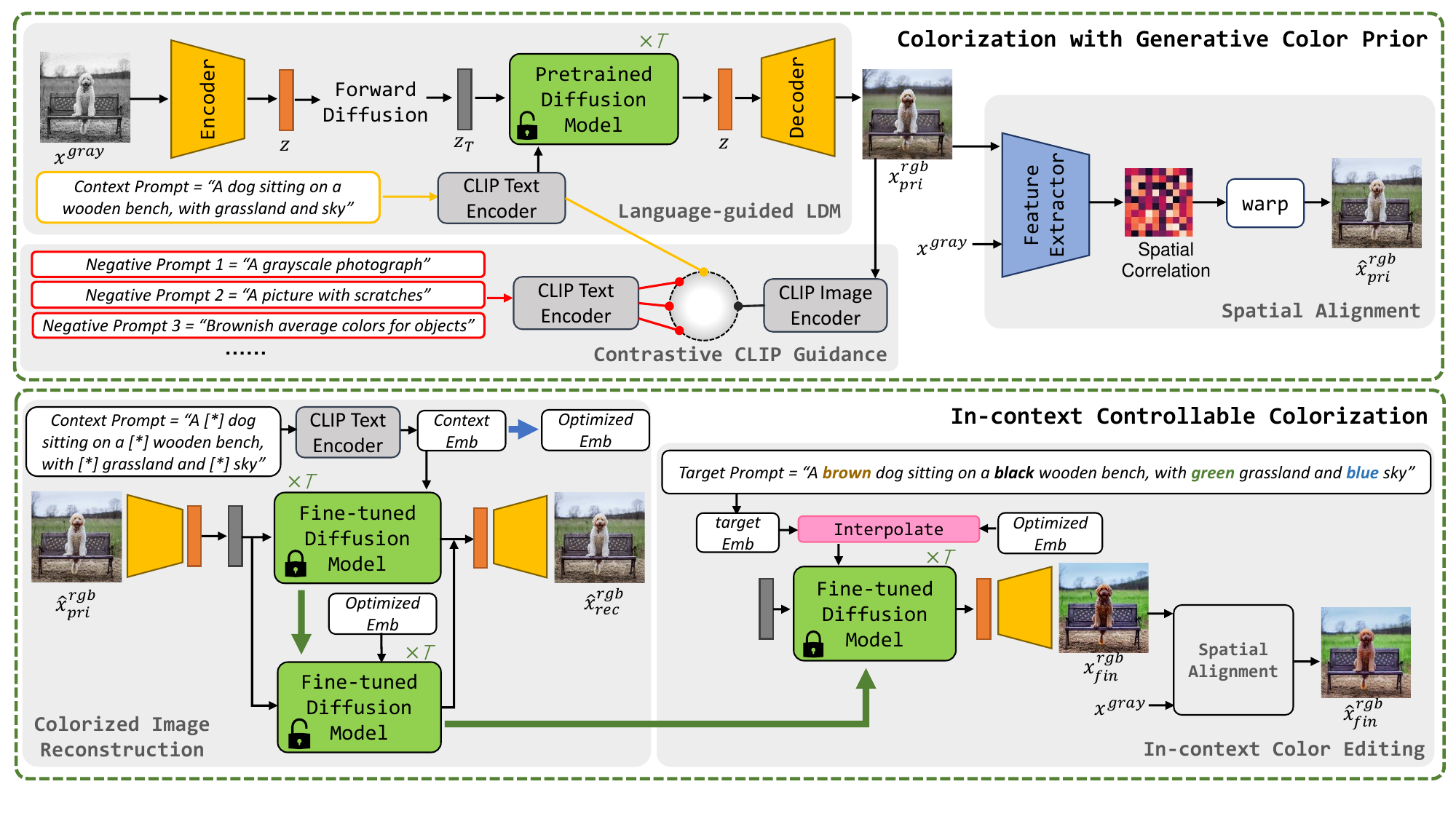}
	\caption{\textbf{Framework of Diffcolor. DiffColor mainly contains two stages: 1) colorization with generative color prior that produces accurate and vivid colorization results; 2) in-context controllable colorization that edits the color of the first-stage output in a way that satisfies the target text prompt. }}  
	\label{fig:framework}
 
\end{figure*}

\section{Approach}

\subsection{Overview architecture}
Our method mainly consists of two stages: \textbf{colorization with generative color prior} and \textbf{in-context controllable colorization}. 
As shown in Figure \ref{fig:framework}, given an input grayscale image $x^{gray} \in \mathbb{R}^{1 \times H \times W}$ and a \emph{context text prompt} $c$ which describes the content of the grayscale image, our goal of the first stage is to produce accurate and vivid colorization results, preserving a maximal amount of details from $x^{gray}$. 
To achieve this, we consider the task of text-guided image colorization through a pre-trained text-to-image latent diffusion model. 
We begin by inputting $x^{gray}$ into a pre-trained encoder, obtaining the corresponding latent code $z$. 
The denoising model $\epsilon_\theta$ takes two inputs, namely the noisy latent $z_T$  and a context text embedding  which is the encoded context text prompt $c$ of $x^{gray}$ by a CLIP text encoder.  
Then, we fine-tune the score function in the reverse diffusion process using a contrastive loss that converts grayscale image to a primary colorized image $x^{rgb}_{pri}$. 
Next, to guarantee the reconstruction of the grayscale image's details, a spatial alignment module is utilized to align the content of the grayscale image $x^{gray}$ and the primary colorized image $x^{rgb}_{pri}$, obtaining colorized image $\hat{x}^{rgb}_{pri}$.

In the second stage, taking the colorized image $\hat{x}^{rgb}_{pri}$ and a \emph{target text prompt} $\hat{c}$ which is constructed by adding color descriptions in front of the object words in the context text prompt $c$ as inputs, our goal is to achieve a fine-tuned model that can edit the color of the image in a way that satisfies the given text while achieving high fidelity. Inspired by DreamBooth \cite{ruiz2022dreambooth} for better image identity capture, we rewrite the context text prompt for fine-tuning like “A [*] dog sitting on a [*] wooden bench.”, where ``[*]'' is a unique identifier. Then we optimize the context text embedding to find the one that maximizes the similarity with the colorized image within the proximity of the target text embedding, as well as fine-tune the diffusion models to improve the alignment between the generated image $\hat{x}^{rgb}_{rec}$ and the colorized image $\hat{x}^{rgb}_{pri}$. In inference time, we perform a linear interpolation between the optimized context text embedding and the target text embedding to obtain a solution that achieves high fidelity to the input image while maintaining semantic alignment with the target text, realizing in-context controllable colorization.


\subsection{Colorization with Generative Color Prior}
In this stage, we aim to leverage rich color priors encapsulated in a pretrained generative model to guide the colorization process. This stage mainly consists of a pre-trained language-guided latent diffusion model, a CLIP guidance, and a spatial alignment module.

\textbf{Language-guided latent diffusion model.} Specifically, the base model for this stage is the latent diffusion model (LDM) \cite{rombach2022high}, which consists of an auto-encoder trained on images and a diffusion model learned on the latent space. The encoder $\mathcal{E}$ of the auto-encoder first encodes the given grayscale image $x^{gray}$ to a latent representation $z$, i.e., $z=\mathcal{E}(x^{gray})$. The diffusion model is trained to produce latent
codes within the pre-trained latent space. The language-guided LDM is learned as follows:

\begin{equation}\label{eqa:ldm}
\mathcal{L}_{\text{LDM}} = \mathbb{E}_{t, \epsilon}[||\epsilon-\epsilon_\theta(z_t, t, \mathcal{E}_{\text{clip}}^{text}(c))||^2_2],
\end{equation}

\noindent where $t$ is the time step, $z_t$ is the latent noised to time $t$, $\epsilon$ is the unscaled noise sample, $\epsilon_\theta$ is the denoising model, and $\mathcal{E}_{\text{clip}}^{text}$ is the CLIP text encoder which encodes the context text prompt $c$ describing the content of
the grayscale image to a context text
embedding.   

\textbf{CLIP guidance.} CLIP \cite{radford2021learning} was proposed to efficiently learn semantic representation from texts and images. It employs a text encoder and an image encoder that are pre-trained to identify the correspondence between texts and images. In this stage, we leverage a pre-trained CLIP model to extract knowledge effectively and build CLIP guidance through a contrastive loss, which makes colorized image $x^{rgb}_{pri}$ (sampled from LDM) more similar to context text $c$ and more dissimilar to other anti-color texts $c^{-}$s which are called negative text prompts, as shown below:

\begin{equation}
\mathcal{L}_{\text{CST}} =-\log \frac{\exp \left(e^{\top} e^{+}\right)}{\exp \left(e^{\top} e^{+}\right)+\sum_{i=1}^{L} \exp \left(e^{\top} e_{i}^-\right)},
\end{equation}

\noindent where  $e=\mathcal{E}_{\text{clip}}^{img}(x^{rgb}_{pri})$, $e^+=\mathcal{E}_{\text{clip}}^{text}(c)$, $e_{i}^-=\mathcal{E}_{\text{clip}}^{text}(c^{-}_i)$,  $\mathcal{E}_{\text{clip}}^{img}$ is the CLIP image encoder, and $L$ is the number of anti-color texts. As shown in Figure \ref{fig:framework}, we can set several anti-color texts, such as ``A grayscale photograph.'', ``A picture with scratches.'' and so on, to prevent colorized image from falling into such situations. The diffusion guidance loss is thus set to the weighted sum $\mathcal{L}_{\text{LDM}}+\lambda\mathcal{L}_{\text{CST}}$.

\textbf{Spatial alignment.} The sampling process of diffusion model is inherently stochastic, meaning that images generated from the diffusion model can not keep the inherent image details, even in the case of a deterministic sampling process like DDIM. Therefore, the reconstruction of the grayscale image's details cannot be guaranteed. Therefore, we utilize a spatial alignment module as previous works \cite{wu2021towards}, 
such as CoCosNet \cite{zhou2021cocosnet}, which is pre-trained by matching grayscale images and their corresponding warped color images, to align primary colorized image $x^{rgb}_{pri}$ and grayscale image $x_{gray}$, obtaining colorized image $\hat{x}^{rgb}_{pri}$.



\subsection{In-context Controllable Colorization}
In the second stage, taking the colorized image $\hat{x}^{rgb}_{pri}$ and a target text prompt $\hat{c}$ as inputs, our goal is to colorize the image in a way that satisfies the given target text while achieving high fidelity. To achieve in-context controllable colorization, we divide the colorization process into two steps as inspired by DreamBooth \cite{ruiz2022dreambooth} and SINE \cite{zhang2022sine}, i.e., colorized image reconstruction and in-context color editing.  

\textbf{Colorized image reconstruction. }For better image identity capture, we rewrite the context text prompt $c$ for fine-tuning like ``A [*] dog sitting on a [*] wooden bench.'' as $c'$, where ``[*]'' is a unique identifier as used in DreamBooth. We separate the reconstruction process into two steps as \cite{kawar2022imagic}. We first optimize the context text embedding of $c'$ to find the one that maximizes the similarity with the colorized image $\hat{x}^{rgb}_{pri}$ within the proximity of the context text embedding. Specifically, we freeze the parameters of the latent diffusion model, and optimize the context text embedding $\mathcal{E}_{\text{clip}}^{text}(c')$ to minimize the  denoising diffusion loss function as Equation \ref{eqa:ldm}, obtaining optimized embedding $e_{opt}$. After that, we fine-tune the diffusion model to further improve the alignment between the new generated image $\hat{x}^{rgb}_{rec}$ and the primary colorized image $\hat{x}^{rgb}_{pri}$, in which we optimize the model parameters using Equation \ref{eqa:ldm} while freezing the optimized embedding $e_{opt}$.

\textbf{In-context color editing.} In inference time, our method can produce different colorization results by modifying target text prompts without any fine-tuning,  and can achieve object-level controllable colorization results, realizing the in-context controllable colorization. Specifically, the target text prompt $\hat{c}$ is constructed by adding color descriptions in front of the object words in the context text prompt $c$, such as ``A brown dog sitting on a purple wooden bench.''. Then, we perform a linear interpolation between the optimized embedding $e_{opt}$ and the target text embedding $e_{tgt}=\mathcal{E}_{\text{clip}}^{text}(\hat{c})$ encoded from the target text prompt $\hat{c}$ to obtain a solution that achieves high fidelity to the input image while maintaining semantic alignment with the target text:
\begin{equation}
\bar{e} = \eta e_{tgt} + (1 - \eta) e_{opt},
\end{equation}
\noindent where $\eta$ is a weight parameter that controls the importance of two embeddings. We then apply the base generative diffusion process using the fine-tuned model, conditioned on $\bar{e}$. This generative process outputs high-resolution colorized image $x^{rgb}_{fin}$. Finally, a spatial
alignment module is also utilized to align the content of $x_{gray}$ and $x^{rgb}_{fin}$ , obtaining our final colorized image $\hat{x}^{rgb}_{fin}$.

\begin{figure*}[!h]
	\centering
	\includegraphics[width=0.98\textwidth]{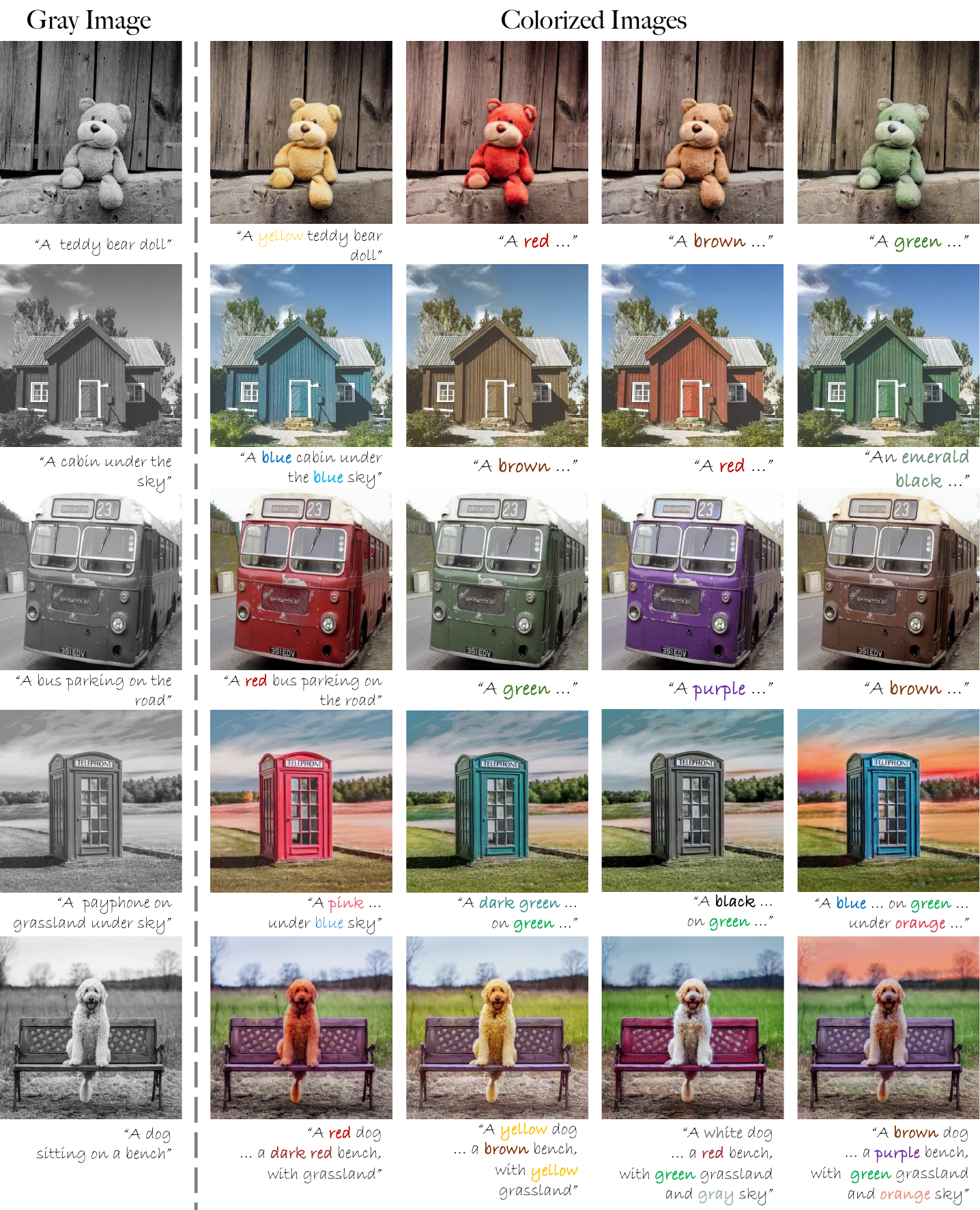}
	\caption{\textbf{Different color prompts applied to the same gray image.}}  
	\label{fig:multi_colors}
\end{figure*}

\section{Experiments}
In this section, we provide both quantitative and qualitative evaluations of our proposed method. We compare our method with state-of-the-art works in both unconditional colorization and text-conditioned colorization respectively. In addition, we perform ablation studies to analyze the impact of different components in our method.
\subsection{Implementation details}
\subsubsection{Network Implementation}
In this paper, we implement our proposed DiffColor with the Stable Diffusion model \cite{rombach2022high} and CoCosNet \cite{zhou2021cocosnet} for 512$\times$512-pixel grayscale images.
Stable Diffusion is a two-stage model which first compresses the image from pixel space into latent space, then perform the noising and denoising process in latent space with less cost. For our 512$\times$512-pixel images, this model applies the diffusion process in a 4$\times$64$\times$64 latent space.
Additionally, to achieve spatial alignment, we simply train a CoCosNet for 20 epochs on 10,000 color and grayscale image pairs without additional model modification.
The implementation details of our proposed DiffColor can be described as:

\textbf{Colorization with Generative Color Prior:} In order to use the prior knowledge of the diffusion model to colorize the given grayscale image $x^{gray}$ according to the context prompt $c$, we use $\mathcal{L}_{\text{LDM}}+\lambda\mathcal{L}_{\text{CST}}$ to make sure that the primary colorized image $x^{rgb}_{pri}$ can be reconstructed and colorized consistent with grayscale image and context prompt. In our experiments, we set $\lambda$ to 0.5 for the optimization process's stability and a more balanced result on reconstruction and colorization. For the fine-tuning of the diffusion model, we freeze the encoder and decoder and only fine-tune the U-Net model for 1500 steps with a learning rate of 2e$-$6. After fine-tuning we can get a colorized image that basically matches the original grayscale image, where the color of the objects comes from the prior knowledge of the diffusion model. Then, with the help of the spatial alignment module, we can get a colorized image $\hat{x}^{rgb}_{pri}$ with the grayscale image's details being preserved.

\textbf{In-context Controllable Colorization:} We rewrite the context prompt $c$ by adding a unique identifier ``[*]'' in front of each object word so that we can better bind the objects in the image to words in the prompt. In order to achieve in-context color editing, we first need to reconstruct the colorized image $\hat{x}^{rgb}_{pri}$. We first optimize the embedding of the rewritten context prompt for 500 steps with a learning rate of 1e$-$3. Then we fine-tune the diffusion model for 1000 steps with a learning rate of 1e$-$6. Finally, we can get a fine-tuned diffusion model that can faithfully reconstruct the colorized image $\hat{x}^{rgb}_{pri}$ based on the optimized embedding $e_{opt}$, which makes subsequent color editing easy to accomplish. To achieve color editing, we first need to get a target prompt, which can be simply obtained by adding different color words in front of each unique identifier of the context prompt. The edited colorized image $\hat{x}^{rgb}_{fin}$ can be obtained by simply performing interpolation between the optimized embedding $e_{opt}$ and the target embedding $e_{tgt}$ with a weight parameter $\eta$. During  experiments, we found that setting the interpolation parameter $\eta$ in [0.7, 1) leads to the best visual effect. 

This whole process takes about 11 minutes on a single A100 GPU.

\subsubsection{Evaluation Metrics}
To assess the overall quality and fidelity of the generated images, we employ several evaluation metrics: Frechet Inception Distance (FID) \cite{heusel2017gans} that quantifies the statistical similarity between the generated images and the ground-truth images; Learned Perceptual Image Patch Similarity (LPIPS) \cite{zhang2018unreasonable} that measures the perceptual similarity between the colorized images and the ground-truth images. In addition, Peak Signal-to-Noise Ratio (PSNR) and  Structural Similarity Index Measure (SSIM) \cite{zhou2004image} are presented for unconditional colorization methods. For text-conditioned colorization, the CLIP score \cite{radford2021learning} is utilized to measure the relevance between the colorized images and the text prompts, evaluating how well the generated images match the provided textual descriptions.

\subsubsection{Datasets}
Since we aim to produce high-fidelity image colorization results and our model does not require large data training, we follow Imagic \cite{kawar2023imagic} and SINE \cite{zhang2022sine} to collect high-resolution, free-to-use images from Unsplash and Flickr for our experiments. Similarly, the dataset we collect has 60 images with various categories, such as dog, cat, car, house, tree, etc.

\subsection{ Qualitative evaluation}

We demonstrate the versatility of our proposed method, DiffColor, by applying it to a range of real-world images with diverse objects. We use simple text prompts that describe different colors of objects in the generated image. We collected high-resolution, free-to-use images from Unsplash and Flickr for our experiments. After optimization, we randomly generate eight results for each grayscale image using different $\eta$ and select the best one. DiffColor can apply various image colorizations via text prompts, including single-object color editing and multiple-object color editing, as shown in Figure \ref{fig:title_page} and Figure \ref{fig:multi_colors}. Moreover, we demonstrate that DiffColor enables post-text color editing by adding new text prompt at the end of the original text prompt, e.g., adding ``with yellow grassland'' at the end of ``A dog sitting on the bench'', as shown in the last row of Figure \ref{fig:multi_colors}. The text prompts specify the desired modifications to different objects in grayscale images, and we can observe that 1) our method can accurately change the colors of different objects without color bleeding while preserving the remaining parts of the image colors; 2) except for specially given colors, our method can recover vivid and natural colors to the original grayscale images due to the powerful diffusion model priors.

\begin{figure*}[!h]
	\centering
	\includegraphics[width=0.98\textwidth]{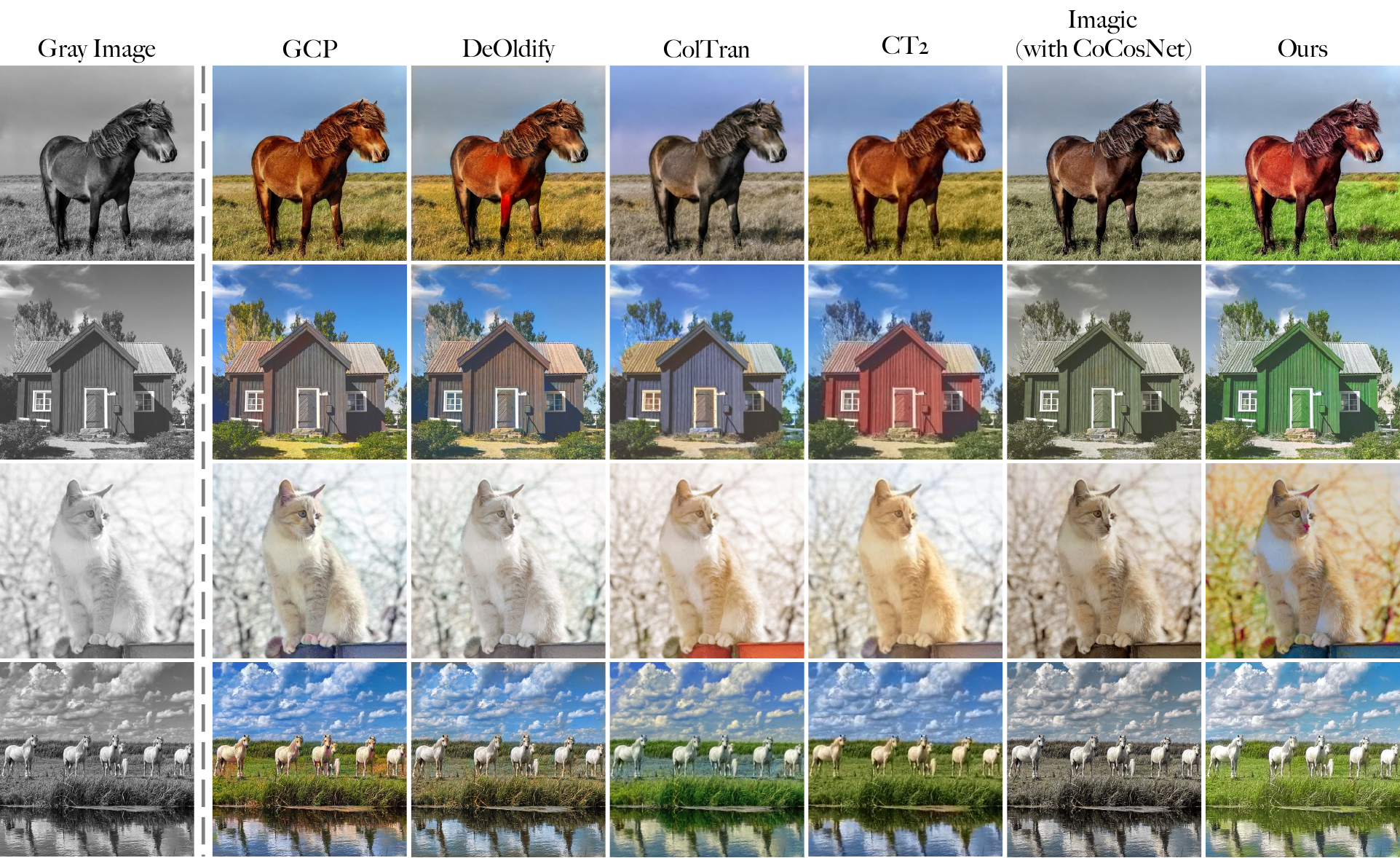}
	\caption{\textbf{Comparison with other existing unconditional image colorization methods.}}  
	\label{fig:comparison}
\end{figure*}

\begin{figure*}[!h]
	\centering
	\includegraphics[width=0.98\textwidth]{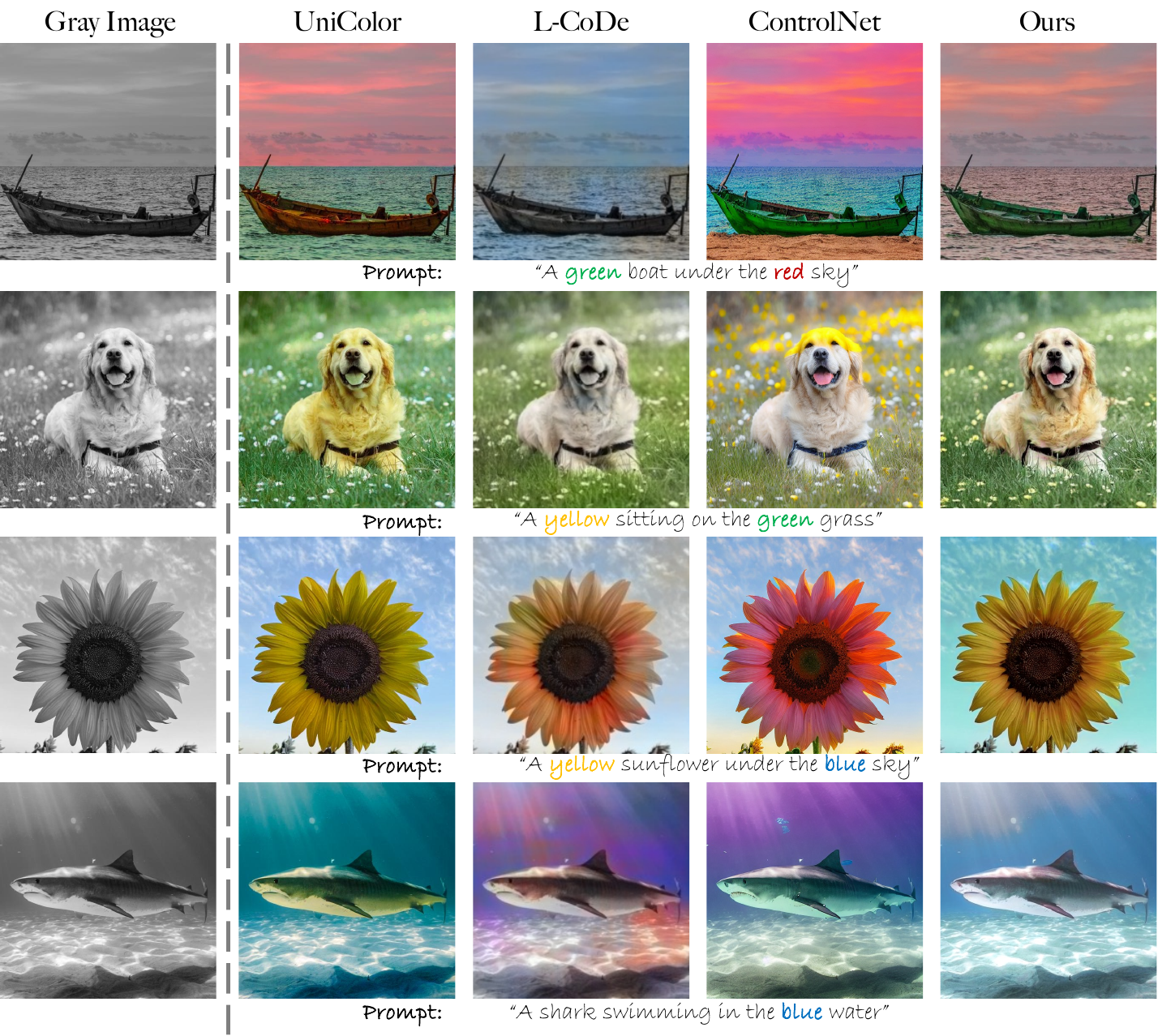}
	\caption{\textbf{Comparison with other existing text-conditioned  image colorization methods.}}  
	\label{fig:comparison2}
\end{figure*}

\subsection{ Comparison to state-of-the-arts}

\subsubsection{Unconditional colorization}
For unconditional colorization, we conduct a comparison with several state-of-the-art (SOTA) techniques capable of colorizing grayscale images without requiring any color hints: namely  GCP \cite{wu2021towards}, DeOldify \cite{salmona2022deoldify}, ColTran \cite{kumar2021colorization}, CT2 \cite{weng2022ct} and Imagic \cite{kawar2023imagic}. We use results from the first stage of our method for comparison in this part. DeOldify \cite{salmona2022deoldify} uses GANs to realistically colorize images and is particularly effective for restoring old images and videos. GCP \cite{wu2021towards} is a two-stage GAN-based framework that generates diverse and vivid colorizations by incorporating multiple color suggestions with grayscale images. ColTran \cite{kumar2021colorization} is a self-supervised transformer model that achieves state-of-the-art performance without relying on handcrafted priors or paired training data.  CT2 \cite{weng2022ct} employs a transformer architecture with color tokens for color representation, leveraging self-attention to efficiently colorize images while preserving their structure and details. In addition, we also compare our method with Imagic \cite{kawar2022imagic}, which also has the capability of recovering image colors. However, since Imagic's generated results may not be perfectly aligned with the original grayscale image, we also apply CoCosNet to spatially align the results for a fair comparison. We evaluate our DiffColor method using the same context prompts without color hints, which are designed by human priors, as Imagic.


\textbf{Qualitative Comparison.} The results of the comparison between our DiffColor and other unconditional methods are displayed in Figure \ref{fig:comparison}. As we can see, our method can produce high-fidelity colorized images with vivid and diverse colors,  without color bleeding. For example, in the second row of Figure \ref{fig:comparison}, when the gray image contains a house with two symmetric roofs, unlike DiffColor, most of  the existing methods tend to colorize the two symmetric roofs with two different colors, lacking the capability of globally semantic perception. In addition, most of the existing methods usually produce images of  brownish average colors in the third row. Although Imagic has a powerful ability to manipulate images' content, it fails to transfer gray images to vivid color images,  which demonstrates the necessity of the proposed CLIP guidance with contrastive loss.

\textbf{Quantitative Comparison.} To better show our re-colored images' quality, we provide quantitative evaluations using FID, PSNR, SSIM and LPIPS metrics. As shown in Table \ref{tbl:comparison_no_clip}, our method still achieves the best performance than other methods in four different metrics, indicating that our method can recover more vivid and natural colors with diffusion model priors.

\subsubsection{Text-conditioned colorization}
For text-conditioned colorization, We conduct a comprehensive comparison of our proposed DiffColor with three state-of-the-art methods, i.e.,  UniColor \cite{huang2022unicolor}, L-CoDe \cite{weng2022code}  and ControlNet \cite{ControlNet}. Both L-CoDe and UniColor are object-level image colorization methods and can realize colorization results with color text prompts. ControlNet proposed a neural network structure to control pretrained large diffusion models to support additional input conditions, for colorization task we use the grayscale images as visual condition and text prompts as text condition. We use results from the second stage of our method for comparison in this part.

\textbf{Qualitative Comparison.} The results of the comparison between our DiffColor and other text-conditioned methods are displayed in Figure \ref{fig:comparison2}. In this part, we present visual quality comparisons involving four distinct samples, each associated with different language descriptions. These descriptions include instances where color words and object words are clearly aligned. From the results, we can find that our proposed method emerges as the superior approach in terms of synthesizing visually pleasing and description-consistent colorization results.


\textbf{Quantitative Comparison.} To better demonstrate our method's ability in the task of text-guided image colorization, we provide quantitative evaluations comparing with state-of-the-art text-guided image colorization methods. For methods that employ text prompts to colorize grayscale images, we place particular emphasis on the CLIP Score \cite{radford2021learning} that measures the consistency between colorized objects and color prompts. As shown in Table \ref{tbl:comparison_clip}, our text-based method obtains the best CLIP similarity, which indicates that our method can accurately locate the objects and colorize them with accurate colors. In addition, it shows that our method can produce higher image quality and fidelity than other methods with the best FID and LPIPS scores.


\begin{table}[t]      
\centering      
\normalsize        
\caption{Quantitative comparison of unconditional image colorization methods.}    

\begin{tabular}{@{}lcccc@{}}      
\toprule      
& FID $\downarrow$  & PSNR $\uparrow$ & SSIM $\uparrow$    & LPIPS $\downarrow$\\ \midrule  
GCP \cite{wu2021towards} & 78.56 & 26.51 & 0.8714 & 0.29 \\
DeOldify \cite{salmona2022deoldify} & 56.63 & 27.97 & 0.8867 & 0.26 \\ 
ColTran \cite{kumar2021colorization} & 73.24 & 27.88 & 0.8833 & 0.27 \\  
CT2 \cite{weng2022ct} & 58.03 & 29.39 & 0.8966 & 0.25 \\  
Imagic \cite{kawar2022imagic} & 121.33 & 22.41 & 0.8023 & 0.35 \\  
Ours-1st & \textbf{48.67} & \textbf{29.83} & \textbf{0.9063} & \textbf{0.22} \\ \bottomrule      
\end{tabular}   
\label{tbl:comparison_no_clip} 
\end{table}

\begin{table}[t]      
\centering      
\normalsize      
    
\caption{Quantitative comparison of text-conditioned image colorization methods.}   
\begin{tabular}{@{}lccc@{}}      
\toprule      
& FID $\downarrow$  & CLIP Score $\uparrow$ & LPIPS $\downarrow$ \\ \midrule  
UniColor \cite{huang2022unicolor} & 62.87 & 28.20 & 0.24 \\  
L-CoDe \cite{weng2022code} & 87.80 & 25.42 & 0.28 \\ 
ControlNet \cite{controlnetpaper} & 65.28 & 27.08 & 0.25 \\  
Ours-2nd & \textbf{51.30} & \textbf{28.62} & \textbf{0.22} \\ \bottomrule      
\end{tabular}    
\label{tbl:comparison_clip}
\end{table}

\begin{table}[t]  
    \centering  
    \small  
    \caption{User study results. Preference rates of unconditional image colorization methods.}  
    \begin{tabular}{@{}cccccc@{}}  
        \toprule  
        GCP    & DeOldify & ColTran & CT2  & Imagic  & Ours-1st   \\  
        \midrule  
         12.3\% &   9.7\%& 16.3\% & 10.7\%& 3.5\% & \textbf{47.5}\% \\  
        \bottomrule  
    \end{tabular}  
    \label{tab:user}  
\end{table}

\begin{table}[t]  
    \centering  
    \small  
    \caption{User study results. Preference rates of text-conditioned image colorization methods in terms of image quality and semantic consistency.}  
    \begin{tabular}{@{}ccccc@{}}  
        \toprule  
        &  UniColor &  L-CoDe   & ControlNet  & Ours-2nd   \\  
        \midrule  
    Quality & 25.6\%  &  13.7\%    & 19.5\% & \textbf{41.2}\% \\  
    Consistency &  27.3\%   &  8.1\%   & 26.5\% & \textbf{38.5}\% \\  
        \bottomrule  
    \end{tabular}  
    \label{tab:user2}  
\end{table}

\subsubsection{User Study.} To assess the subjective quality of our method in terms of color vividness, colorfulness, and consistency, we conducted a user study comparing it with other colorization methods. 

For unconditional colorization, we recruited 25 participants with normal or corrected-to-normal vision and without color blindness. We selected 30 grayscale images from various categories of image content. For each image, we displayed the grayscale version on the leftmost side, while five colorization results were displayed randomly to avoid potential bias. The participants were asked to select the best-colorized image based on the quality of colorization in terms of vividness and colorfulness. We present the results in  Table \ref{tab:user}. Our DiffColor method was significantly preferred (47.5\%) by the users compared to other colorization methods, demonstrating a distinct advantage in producing natural and vivid results.

For text-conditioned colorization, we recruited another 25 participants. In the evaluation process, except for selecting the best-colorized image in terms of vividness and colorfulness, participants are presented with a caption describing a color image, and they are asked to choose the colorized image that is most consistent with the caption. As shown in Table \ref{tab:user2}, our approach consistently outperforms the other methods, achieving the highest scores in both scenarios, i.e., image quality and semantic consistency.

\begin{figure*}[t]
	\centering
	\includegraphics[width=1\textwidth]{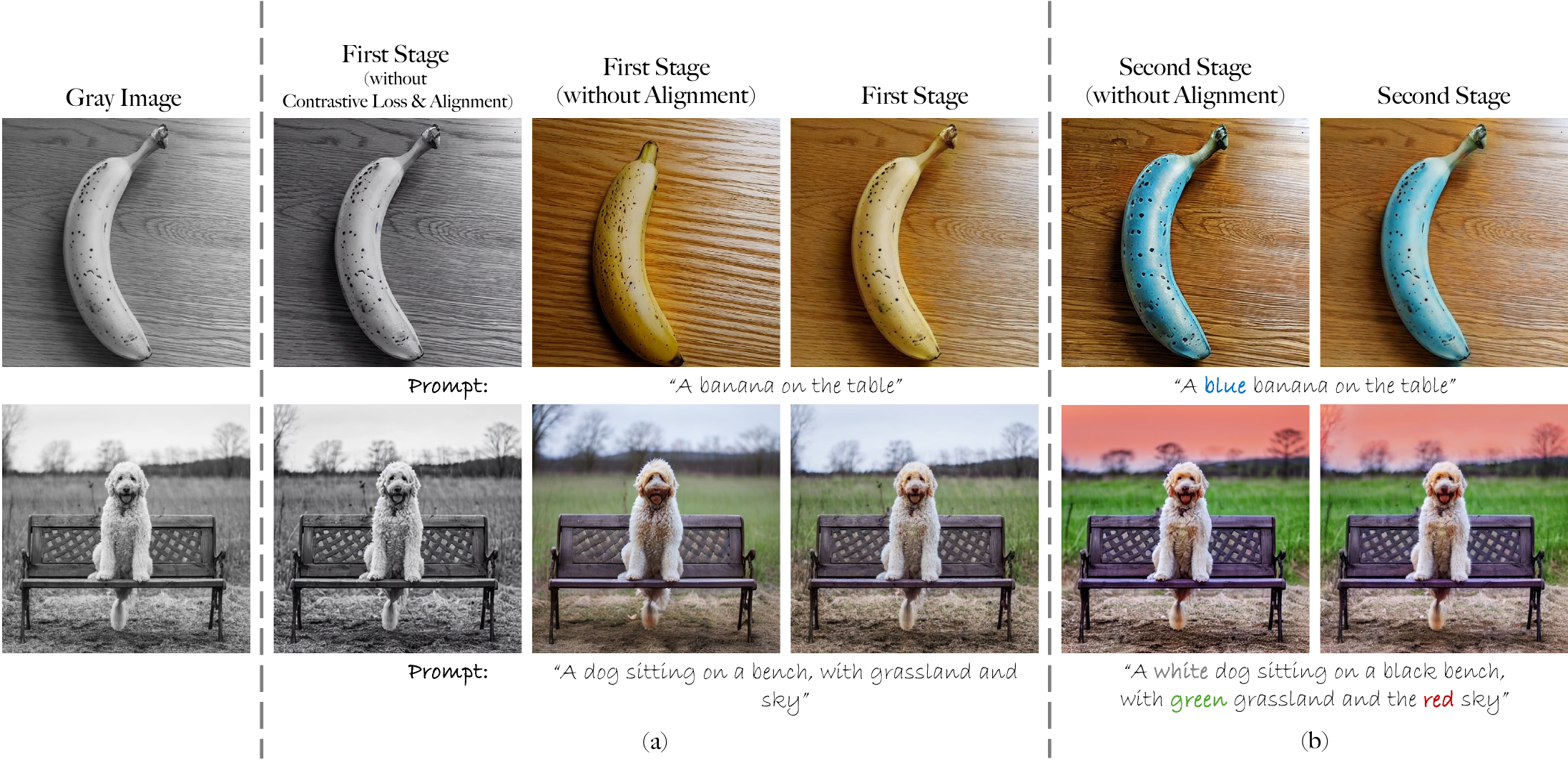}
	\caption{\textbf{Ablation study on DiffColor's each module.}}  
	\label{fig:ablation}
\end{figure*}
\begin{figure*}[t]
	\centering
	\includegraphics[width=0.8\textwidth]{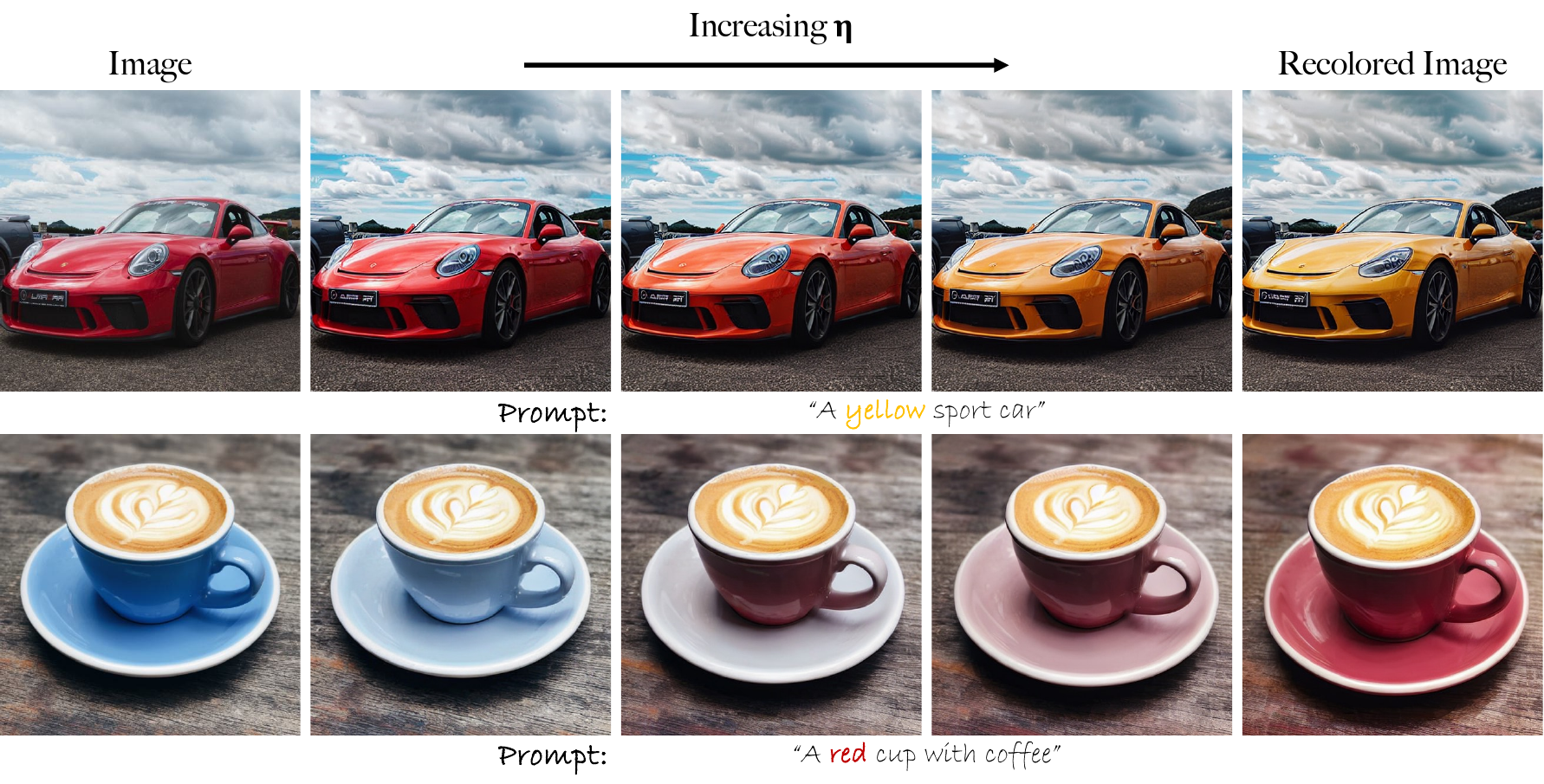}
	\caption{\textbf{Ablation study on interpolation parameter $\eta$. In this figure, we directly input real color images to the second stage of our method.}}  
	\label{fig:ablation2}
\end{figure*}
\subsection{Ablation study}
\textbf{Colorization with Generative Color Prior.}
 We exhibit the initial colorization results obtained from the grayscale image after processing solely through the first stage as shown in Figure \ref{fig:ablation} (a). By incorporating the contrastive loss derived from the CLIP, we effectively confine the entire output of the first stage within a color space. Furthermore, the reconstruction loss allows us to concurrently consider the quality of the image reconstruction. Consequently, through capitalizing on the extensive prior knowledge embedded within the diffusion model, we successfully achieve grayscale image colorization based exclusively on content cues. It is important to note that our objective during the first stage does not entail the diffusion model's ability to perfectly reconstruct the grayscale image's details while simultaneously performing colorization. Instead, we permit a certain degree of detail loss, which is subsequently mitigated through the utilization of spatial alignment techniques. 
 
\textbf{Controllable Colorization.}
 We showcase the color editing outcomes of the second stage by applying it directly to real color images as shown in Figure \ref{fig:ablation2}. This demonstrates the second stage's versatility, as it is not only capable of functioning effectively within our proposed two-stage methodology but can also operate independently as a semantic color alteration technique for color images.
 
\textbf{Spatial Alignment.}
We compare the results with and without the spatial alignment as shown in Figure \ref{fig:ablation}. It turns out that, although the output of the diffusion model is not impeccable in terms of reconstruction, it is generally consistent with the provided color prompt. By employing CoCosNet to achieve spatial alignment between the output and the grayscale image, we manage to extract the colors from the diffusion model's direct output while maintaining the structure and details inherent in the original image. As shown in Table \ref{tbl:ablation_study}, the alignment module significantly improves image structure preservation in the first stage, and our method can produce results with a highly similar image structure in the second stage.  

\begin{table}[t]  
\centering  
\caption{Quantitative results for ablation study. }  
\label{tbl:ablation_study}  
\small  
\begin{tabularx}{\columnwidth}{@{}lXXXX@{}}  
\toprule  
Stage & \multicolumn{2}{c}{First} & \multicolumn{2}{c}{Second} \\ \cmidrule(lr){2-3} \cmidrule(lr){4-5}  
& W/O Align. & W/ Align. & W/O Align. & W/ Align. \\ \midrule  
PSNR & 21.13 & 29.83 & 25.56 & 29.16 \\  
SSIM & 0.7643 & 0.9063 & 0.8476 & 0.8934 \\ \bottomrule  
\end{tabularx}  
\end{table}

\textbf{Interpolation Parameter $\eta$.} Figure \ref{fig:ablation2} displays several different results with increasing $\eta$ for the same image. This approach provides the user an option to colorize gray images when the color description is ambiguous or imprecise.

\section{Conclusion}

In this paper, we presented a novel method called DiffColor that addresses the limitations of classic and reference-based image colorization methods. Our approach leverages the power of pre-trained diffusion models to produce vivid and diverse colors conditioned on a text prompt without requiring any additional inputs. DiffColor consists of two main stages: colorization with generative color prior and in-context controllable colorization. We fine-tune a pre-trained text-to-image model using a CLIP-based contrastive loss to generate colorized images and optimize the text embedding to align the colorized image with the target text prompt. Moreover, we fine-tune a diffusion model to enable high-quality image reconstruction. Our method allows for in-context colorization and achieves object-level controllable colorization results.

Extensive experiments and user studies demonstrate that DiffColor outperforms previous works in terms of visual quality, color fidelity, and semantic consistency. We believe that our approach represents a significant advancement in text-guided image colorization using diffusion models and provides a convenient and powerful tool for producing high-quality colorized images based on textual descriptions. Our method has the potential to benefit a wide range of applications, including art, fashion, and product design, among others. Further research could explore the use of DiffColor for other image editing tasks beyond colorization.

\bibliographystyle{IEEEtran}
\bibliography{ref}

\end{document}